\documentclass{article}

\usepackage{microtype}
\usepackage{graphicx}
\usepackage{subfigure}
\usepackage{booktabs} 
\usepackage{optidef}
\usepackage{amsmath}
\usepackage{dsfont}
\DeclareMathOperator*{\argmax}{arg\,max}

\usepackage{float}

\usepackage{hyperref}

\usepackage[accepted]{icml2020}

\begin{document}
\twocolumn[
\icmltitle{On Detecting Data Pollution Attacks On Recommender Systems Using Sequential GANs}


\icmlsetsymbol{equal}{*}

\begin{icmlauthorlist}
\icmlauthor{Behzad Shahrasbi}{equal,wm}{}
\icmlauthor{Venugopal Mani}{equal,wm}{}
\icmlauthor{Apoorv Reddy Arrabothu}{equal,wm}{}
\icmlauthor{Deepthi Sharma}{wm}{}
\icmlauthor{Kannan Achan}{wm}{}
\icmlauthor{Sushant Kumar}{wm}{}
\end{icmlauthorlist}

\icmlaffiliation{wm}{WalmartLabs}
\icmlcorrespondingauthor{Behzad Shahrasbi}{behzad.shahrasbi@walmartlabs.com}
\icmlkeywords{Recommender Systems, Data Pollution Attack, GANs}
\vskip 0.3in
]

\printAffiliationsAndNotice{\icmlEqualContribution}

\renewcommand\abstractname{\textsc{ABSTRACT}}
\begin{abstract}
Recommender systems are an essential part of any e-commerce platform. Recommendations are typically generated by aggregating large amounts of user data. A malicious actor may be motivated to sway the output of such recommender systems by injecting malicious datapoints to leverage the system for financial gain. In this work, we propose a semi-supervised  attack detection algorithm to identify the malicious datapoints. We do this by leveraging a portion of the dataset that has a lower chance of being polluted to learn the distribution of genuine datapoints. Our proposed approach modifies the Generative Adversarial Network architecture to take into account the contextual information from user activity. This allows the model to distinguish legitimate datapoints from the injected ones.
\end{abstract}

\section{INTRODUCTION}
Recommender systems play a key role in driving user engagement on many e-commerce platforms. They are typically used to provide personalized item recommendations to the users. The goal of recommender systems is to increase revenue and profitability for merchants by showing relevant and novel items to the users on e-commerce platforms \cite{aggarwal2016recommender}. Given the scale of e-commerce revenues, a malicious actor can obtain significant financial gains by altering the output of these recommender systems. Many recommender systems are trained on user generated data such as co-views and co-purchases. Figure 1 shows how a malicious actor can introduce spurious user engagement events to inject polluted data into a recommender system and eventually alter the recommendations. In recent years more sophisticated attack strategies are developed to make fake data points as close as possible to the real data \cite{christakopoulou2019adversarial, fang2020influence}. 
Data pollution attacks in turn can be categorized in two categories, namely, \emph{push attacks} in which the attacker tries to promote one or more target items on the e-commerce platform and \emph{nuke attacks} in which the attacker targets some items and tries to suppress them in the output of the recommender system \cite{chirita2005preventing, gunes2014shilling}.

\begin{figure}[h]
  \centering
  \includegraphics[width=\linewidth]{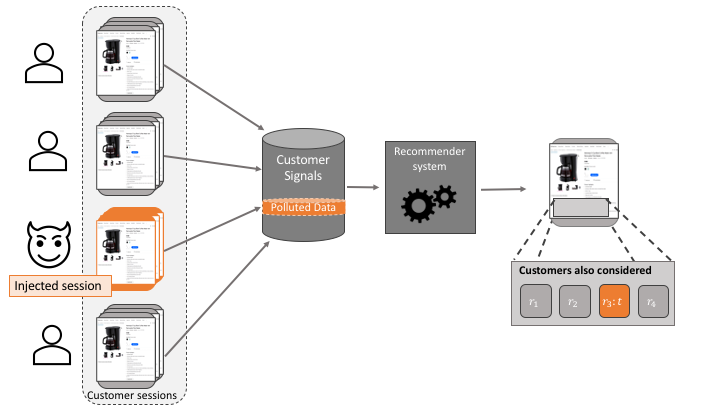}
  \caption{\textbf{Data Pollution Attack Architecture}}
  \label{arch_fig}
\end{figure}
\vspace {-2 mm}

In this work, we consider a scenario in which the attacker injects malicious datapoints in order to increase the visibility of certain target items in the output of the recommender system. As shown in Figure 1, an attacker creates a user engagement session (e.g., browsing items), which includes a target item and other popular items. The data generated from all users' sessions (including the attacker's) gets captured in form of low-level co-view and co-purchase features which are used by the recommender system to train and generate relevant item recommendations. If enough malicious datapoints are injected by the attacker then the target item will appear in more item recommendations than it would normally appear prior to the attack.
We use a standard collaborative filtering based recommender system for our experiments. We demonstrate two types of attacks on this recommender system: (1) White-box attack: where attacker has full knowledge of data distribution of recommender system's training data, and (2) Gray-box attack: where attacker has partial knowledge of the recommender system. Goal of the proposed approach is to identify maliciously injected datapoints in both types of attacks on the recommender system.

\subsection{Related Work}
\cite{xing2013take, gunes2014shilling} surveyed a wide range of data pollution attacks on recommender systems, establishing the breadth of this problem. More recently, \cite{yang2017fake} demonstrated successful fake co-view injection attacks on multiple e-commerce platforms (e.g. eBay, Yelp) utilizing their user-item recommender systems architecture. \cite{christakopoulou2019adversarial, fang2020influence} developed an adaptive attack which tried avoding detection of fake profiles and injected data points. In \cite{christakopoulou2019adversarial}, an adversarial training strategy is developed with the goal to find a user-item matrix, to minimize distance of interaction distributions, and the adversary’s intent is accomplished. 

\cite{hurley_statistical_detection} proposed statistical methods for identifying attacks on recommender systems. \cite{shilling_time_series} and \cite{kalman_filter} attempt to detect attacks through time series anomalies. However, these works focus on explicit user ratings while the e-commerce setting has an implicit nature that is not as temporal. 

Generative Adversarial Networks (GANs) are very effective in learning the true distribution of data and generating realistic datapoints \cite{goodfellow2014generative}. The GAN however, is not suitable to work with discrete sequential data \cite{goodfellow2014generative,yu2017seqgan,gumbel-gan} because the sampling step at the generator's output has undefined gradients. Thus, the application of these architectures in the attack detection literature has been very limited so far because of inherent sequential nature of pollution attacks.  However, \cite{yu2017seqgan,gumbel-gan} addressed this shortcoming of GANs and introduced architectures that are well suited for discrete sequential data (i.e. seqGAN). \cite{di2019survey} surveyed the application of GANs in anomaly detection application. While the survey article did not specifically focus on attack detection, some of the architectures discussed in that paper can be used to detect attacks. For instance \cite{zheng2019one} used the GANs with LSTM-autoencoders for fraud detection in a two step approach. 
\subsection{Contributions}
In this paper we propose a semi-supervised data pollution attack detection strategy. 
The contributions of this paper are the following. First, we introduce a sequential GAN architecture for data pollution attack detection. The novelty of this approach is including the contextual information from user engagement sequences in the attack detection algorithm.
Secondly, we show the effectiveness of the proposed attack detection algorithm under gray-box and white-box attack scenarios on two real world e-commerce datasets.   

\section{SYSTEM MODEL} \label{sysmodel}
In this Section, we explain the components of underlying recommender system and the proposed attack detection architecture. First, we introduce the following notations.

Let, $u_i \in U$ be the $i^{th}$ \emph{user} of the e-commerce platform $\forall i \in \{1, \ldots, |U|\}$. Additionally, $u_i$ is associated with arbitrary number of user sessions $s_{ij} \in S_i$, where $j > 0$ is the $j^{th}$ session of the user $u_i$, $S_i \in \mathbf{S}$, $S_i$ is the set of all sessions of user $u_i$ and $\mathbf{S}$ is the set of all user sessions $S_i$. A user session $s$ is a sequence of user signals $v \in V$ over a span of time $t$ (e.g. a sequence of page browse activities within one session or basket of items that the user purchased). Therefore, a user session $s$ is defined as $s = [v_1, \ldots, v_l \ldots v_{k_{ij}}]$, where $v_l$ represents the $l^{th}$ item that the user interacted with in session $s$. Without loss of generality we assume that the length $k_{ij} <= K$ for any user session sequence $s_{ij}$.

In addition, each item has contextual features (e.g., title, description, brand etc). We can map each item into an embedding space using a language model on items contextual features. In turn a user session can be represented by a sequence of embeddings. 
The item embedding $\boldsymbol{e}_i \in \mathds{R}^d$ for item $v_i$ is
\vskip -0.3in
\begin{equation}\label{eq:1}
    \boldsymbol{e}_i = \mathcal{E}(\mathcal{D}(v_i)),
\end{equation}

where $\mathcal{D}(.)$ is the function that returns contextual features (such as title, description, brand, etc) for the item $v_i$ and $\mathcal{E}(.)$ is the embedding function which generates the $d$ dimensional item embedding from the semantic item features using a pre-trained language model (e.g. Doc2Vec\cite{le2014distributed} or BERT\cite{devlin2018bert}). Accordingly, $\boldsymbol{s_{ij}} = [\boldsymbol{e}_1, \boldsymbol{e}_2, \ldots, \boldsymbol{e}_{k_{ij}}]$.

\subsection{Recommender System}\label{recsys}
In this work, we assume every item page has a set of relevant item recommendation that will be shown to the user below the main item.
Various types of recommender systems such as collaborative filtering approaches \cite{mfsys} and graph based approaches \cite{fang2018poisoning} are shown to be vulnerable to data pollution attacks from attackers with near complete knowledge of the system. In \cite{mobasher2005influence}, item-based collaborative filtering was found to more robust to data pollution attacks than user-based collaborative filtering and hence, our focus was on attacking an item-based collaborative filtering. 
The rating matrix is constructed from the set of users $U$ and the set of all items $V$  and used in a standard item-based collaborative filtering recommender. Entries of the rating matrix are \emph{implicit} signals (user engagement signals like views, transactions) rather than explicit ratings. 

From the pool of signals, let $q_{i_{u}}$ be the implicit rating (computed as a function of the signals) of the $i^{th}$ item provided by user $u$ (The rating is zero if the user-item pair has not interacted). The item similarity, $CS(v_i, v_j)$, between item $v_i$ and item $v_j$ is computed as per Equation \ref{cosine_sim}. 
\small
\begin{equation}\label{cosine_sim}
    CS(v_i,v_j) = \frac{\sum_{u\in U} q_{i_u} q_{j_u}}{\sqrt{\sum_{u \in U} q_{i_u}^{2}} \sqrt{\sum_{u \in U} q_{j_u}^{2}}}
\end{equation}
\normalsize
Recommendations are then computed for a given user based on computing the most similar items to the ones that a given user has already rated. Each user is assigned $r$ recommendations (in an application setting, this would be the number of items visible to the user).

With this logic in place, each user gets top $r$ recommendations which is a combination of their own activity and the activity of other users. It follows that to attack such a system, an attacker can detect patterns which are in some way related to the user's activity. This will be discussed further in Section \ref{util}.

\subsection{Sequential GAN Architecture for Data Pollution Attack Detection}
In this section, we introduce our proposed contextual sequential GAN architecture to detect data pollution attacks on recommender systems.
To build the attack detection framework, we take advantage of the seqGAN architecture\cite{yu2017seqgan} to simultaneously learn the distribution of items in user sessions and develop ability to distinguish between the real looking sessions and polluted sessions.
\subsubsection{Semi-supervised learning}
It is not feasible to have access to a labeled dataset with samples from both clean and injected sessions in a real world e-commerce platform, as most attacks go undetected. On the other hand, without having access to labeled data reduces the attack detection problem to anomaly detection problem. In this work, we consider a more realistic scenario in which the e-commerce provider has heuristic knowledge about the user data. Using this heuristic knowledge, the user data can be broken up into \emph{clean} and \emph{polluted} parts. For example, the clean dataset can include sessions that are generated by only logged-in highly engaged users, in which the retailer has high confidence of them being genuine users. While the rest of the sessions that don't fit to the clean part will form the polluted dataset.
Our objective is to train the GAN architecture to learn the distribution of the user signals using clean dataset and then look for datapoints in the polluted set that are unlikely to come from this distribution. 

\subsubsection{Adversarial Architecture for Detection}
\begin{figure}[h]
\centering
\includegraphics[width=\linewidth]{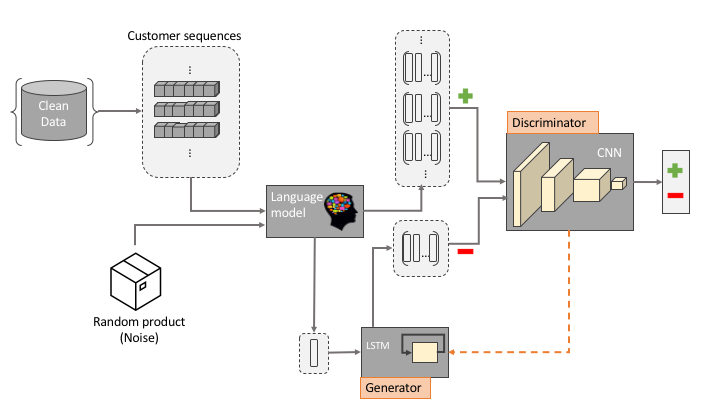}
\caption{\textbf{Proposed sequential GAN architecture for Data pollution attack detection}}
\label{gan_fig}
\end{figure}

In our proposed attack detection architecture, we use the \emph{clean} part of the data to generate the embedding sequences ($\boldsymbol{s_{ij}}$) from user sessions using Equation \ref{eq:1} with Doc2Vec language model. The adversarial model training is started in the same way as presented in seqGAN paper \cite{yu2017seqgan}. The architecture of our proposed sequential GAN for data pollution attack detection is depicted in Figure \ref{gan_fig}.

The proposed algorithm addresses two drawbacks of seqGAN in dealing with high dimensional e-commerce data. In a typical e-commerce platform the number of items can be in order of several hundred thousands to millions. Therefore, applying seqGAN directly to this problem will be impractical because of high dimensionality and extremely high computational cost.


The generative model is an LSTM architecture \cite{hochreiter1997lstm} that maps the item embedding in a session $\boldsymbol{e_1}, \boldsymbol{e_2}, \ldots, \boldsymbol{e_K}$ to a sequence of hidden states denoted by $\boldsymbol{o_1}, \boldsymbol{o_2}, \ldots, \boldsymbol{o_K}$. Thus,  similar to \cite{yu2017seqgan}, the update function $G_{LSTM}(.)$ for the generative model is the following
\begin{equation}
    \boldsymbol{o_l} = G_{LSTM}(\boldsymbol{o_{l-1}},\boldsymbol{e_l})
\end{equation}
$\forall l \in {1, \ldots, K}$. Accordingly the probability distribution of $l^{th}$ item in a session, $y_l$ is determined as follows 
\small
\begin{equation}\label{generator_dist}
    p(y_l|\boldsymbol{e_1}, \boldsymbol{e_2}, \ldots, \boldsymbol{e_k}) = z(\boldsymbol{o_l}) = z(G_{LSTM}(\boldsymbol{o_{l-1}},\boldsymbol{e_{l}}))
\end{equation}
\normalsize
where $z(.)$ is the softmax function. In training step, we do not allow the gradients to update the item embeddings $\boldsymbol{e_l}$s. Thus the contextual item embeddings remain unchanged during pretraining and training steps of the GAN.


During the GAN training, the generator converges to the distribution of real sequences, while the discriminator learns a tight decision boundary around the distribution of real looking sequences. At this point, we can evaluate the \emph{polluted} part of the data using this sequential GAN model. The discriminator that is adversarially trained on clean data learns a very tight decision boundary around the clean sequences and will detect with high success rate even the sophisticated data pollution attack. We evaluate the performance of this detection algorithm on two real world datasets in Section \ref{results}. 

\section{ATTACKER UTILITY} \label{util}
The key part of any good data pollution attack is the utility function of the attacker. Quantifying what constitutes a good data pollution attack is largely outcome based and since our focus in this work is on a push attack, we use the hit-ratio of the target item as in \cite{fang2018poisoning}. The hit-ratio $h_t(r)$ of a target item $t \in V$ that the attacker is promoting can be defined as the fraction of normal users whose top $r$ recommendations feature the injected target item $t$.

\indent Let $U^n \subset U$ be the set of normal users and $S^n \subset S$ be the set of associated sessions for these normal users. The attacker aims to  inject $|U^m|$ malicious users where $|U^m|$ is some percentage $\alpha$ of the $|U^n|$ present in the system. Let $o$ be the maximum number of sessions allowed per user.
Let $S^m_j = \{ s^{m}_{j_1} \ldots s^{m}_{j_o}\}$ be the set of sessions of malicious user $u^{m}_{j}$. The set of associated malicious sessions of the $U^m$ users is then denoted by $\mathbf{S^m} = \{S^{m}_{1}, \ldots, S^{m}_{|U^m|}\}$ , where $|S^m|$ (= $ |U^m| \times o$). Let $U^{n}_{t} \subset U^n$  be the set of normal users for whom the top $r$ recommendation contains the artificially promoted item $t$. The hit-ratio is thus defined as $h_t(r) = \frac{|U^{n}_{t}|}{|U^n|}$.

The optimization problem for the attacker is finding the best user sessions $\mathbf{S^{m*}}$ to maximize $|U^{n}_{t}|$. The optimization is represented in the following equation. 
\small
\begin{argmaxi}  |l|[3]
{\mathbf{S^m}}{h_t(r)
}
{}{\mathbf{S^{m*}} = }
\addConstraint{|U^m|\times o <= \mathrm{budget}}
\addConstraint{\forall s \in \mathbf{S^m} , |s| <= \sigma}{} 
\label{eq4} \end{argmaxi}
\normalsize
The first constraint for the attacker is the budget that is available to them. The budget directly determines the number of users that the attacker can inject into the system. The second constraint is one of session length ($\sigma$). The attacker is restricted to a max session length for two reasons: the session length also has an impact on the budget of the attacker and the attacker should keep the sessions as realistic as possible to escape being pruned away by basic preprocessing steps that the platform might employ before training.

\subsection{White-Box Attack}
The white-box attack requires the attacker to have full knowledge of the data distribution $D$ which includes the user-item ratings matrix $M$ and the knowledge of how the e-commerce providers' recommender system works. It is important to know the vulnerability of the system even in the event of an attacker having full knowledge, as per Kerckhoff’s principle. 

\indent Since the attacker is attempting to maximize hit-ratio of the target item $t$, it is logical to mimic the ratings of the best hit-ratio items $B \subset V$ from the existing recommender system $R$, while trying to avoid detection. The white-box routine is described in Algorithm \ref{white_box_algo}.

\vspace*{-3mm}
\begin{algorithm}[tb]
   \caption{White Box Algorithm}
   \label{white_box_algo}
\begin{algorithmic}
   \STATE {\bfseries Input:} Original Data $D$, Target Item $t$, User Budget $|U^m|$, Session Threshold $o$, Session Length $\sigma$ 
    \STATE {\bfseries Output:} Malicious Matrix $M'$\\
    \textbf{Procedure : }WHITE-BOX ATTACK $D$,$t$,$|U^m|$,$o$,$\sigma$\\
    ${M \leftarrow}$ Form original rating matrix using ${D}$\\
    ${R \leftarrow}$ Generate Recommender System for original users\\
    ${B\leftarrow}$ Set of items with highest hit-ratios in R\\
    ${\mathbf{\overline{M}} \gets \{\O \}}$
   
   \FOR{\texttt{$b \in B$}}
        \STATE ${{I_b}\leftarrow}$ Get user-item interaction dist. of ${b}$, using ${D}$\\
        \FOR{\texttt{$j$ in range $1$ to $|U^m|\times o$}}
            \STATE ${i_{bj} \sim I_b \gets}$ Sample of length ${\sigma}$ from ${I_b}$\\
            \STATE ${i_{bjt} \leftarrow}$ Replace random item in ${i_{bj}}$ with ${t}$ ${\forall}$ $i_{bj}$\\
            \STATE ${\mathbf{S^{m}_{bt}} \leftarrow }$Append ${i_{bjt}}$ to malicious sessions set ${\mathbf{S^{m}_{bt}}}$\\
        \ENDFOR
        \STATE ${M^{*}_{bt} \leftarrow}$ Candidate rating matrix from ${D}$ and ${\mathbf{S^{m}_{bt}}}$ \\
        \STATE ${\mathbf{\overline{M}} \leftarrow}$ Append ${M^{*}_{bt}}$ to set of candidate matrice\\
   \ENDFOR
    \STATE ${M' \leftarrow \argmax_{M^{*}_{bt} \in \mathbf{\overline{M}}} (P(M^{*}_{bt}|M)}$\\
    \textbf{return} ${M'}$\\
\end{algorithmic}
\end{algorithm}

\subsection{Gray-Box Attack}
The second type of attack is a gray-box setting where the attacker has partial knowledge of the data distribution only and no knowledge of the Recommender System. Partial knowledge here is defined as the attacker knowing $p\%$ of the distribution of the original data. Therefore original data distribution $D$ from the white-box setting becomes a partially visible $\widehat{D}$ , from which the approximate rating matrix $\widehat{M}$ is computed.  Since the attacker does not have access to the recommendation algorithm, the attacker uses $\epsilon$ fraction of the budget to normally interact with the platform. On doing this, the attacker gathers the recommendations of these $\epsilon |U^m|$ users to form $\widehat R$ which is an approximation of $R$. Following this, the attacker calculates $\widehat B$ is the set of items with the best hit-ratio in $\widehat R$. The rest of the gray-box procedure \ref{gray_box_algo} follows similar steps to the white-box procedure using $\widehat{B}$ as the set of reference items to generate malicious sessions for $(1-\epsilon)|U_m|$ (which is the remaining budget after the warm up sessions) users having $o$ sessions of length $\sigma$. 

\vspace*{-3mm}

\begin{algorithm}[tb]
   \caption{Gray-Box Algorithm}
   \label{gray_box_algo}
    \begin{algorithmic}
       \STATE {\bfseries Input:} Partial Data $\widehat{D}$, Target Item $t$, User Budget $|U_m|$,Session Threshold $o$, Session Length $\sigma$ , Warm Up Ratio $\epsilon$ \\
        \STATE {\bfseries Output:} Malicious Matrix $M'$\\
        \textbf{Procedure : }GREY-BOX ATTACK $\widehat{D}$,$t$,$|U^m|$,$o$,$\sigma$, $\epsilon$\\
        \STATE $\epsilon * |U^m|$ malicious users warm up the system through regular interaction\\
        \STATE $\widehat{M} \gets$ Form approximate original rating matrix using $\widehat{D}$ and ratings from Line 2\\
        \STATE  $\widehat{R} \gets$ Recommender System Generated for the users from Line 2\\
        \STATE $\widehat{B}\gets$Set of items with highest hit-ratios in $\widehat{R}$\\
        \STATE $\mathbf{\overline{M}} \gets \{\O \}$\\
        \FOR{\texttt{$b \in \widehat{B}$}}
            \STATE $\widehat{I_b} \gets$Get (approx.) interaction distribution of $b$, using $\widehat{D}$\\
            \FOR{\texttt{$j$ in range $1$ to $(1-\epsilon)|U^m|\times o$  }}
                \STATE $\widehat{i_{bj}} \sim \widehat{I_b} \gets$ Sample of length $\sigma$ from $\widehat{I_b}$\\
                \STATE $\widehat{i_{bjt}} \gets$ Replace random item in $\widehat{i_{bj}}$ with $t$ $\forall$ $\widehat{i_{bj}}$\\
                \STATE  $\mathbf{\widehat{S^{m}_{bt}}}\gets$ Append $\widehat{i_{bjt}}$ to malicious sessions set $\mathbf{\widehat{S^{m}_{bt}}}$\\
            \ENDFOR
            \STATE $M^{*}_{bt} \gets$ Candidate rating matrix from $\widehat{D}$ and $\mathbf{\widehat{S^{m}_{bt}}}$\\
            \STATE $\mathbf{\overline{M}} \gets$ Append $\widehat{M^{*}_{bt}}$ to set of candidate matrices\\
        \ENDFOR
        \STATE  $M'\gets \argmax_{M^{*}_{bt} \in \mathbf{\overline{M}}} (P(M^{*}_{bt}|\widehat{M})$  \\
        \STATE \textbf{return} $M'$
    \end{algorithmic}   
\end{algorithm}

\section{EXPERIMENTAL RESULTS} \label{results}
In this section, we evaluate the performance of our proposed proposed detection algorithm using simulated data pollution on two real world e-commerce datasets.
\subsection{Datasets}
For evaluation, we use the following two datasets. First dataset is the open source Instacart grocery data \cite{InstacartData2017}. Instacart grocery dataset has $3.4M$ orders from $206K$ users. The item vocabulary is comprised of $50K$ items. For this experiment, we create one user session from each order. 
We split the user $50/50$ into clean and polluted sections uniformly at random.  


The second dataset is a proprietary dataset from a real world e-commerce platform. The dataset consists of sequences of user view sessions on all items in the category of \emph{electronic products} across a given period of time. A view session includes all the consecutive item pages that a user visits in one browsing session. The curated dataset includes $1.85M$ view sessions from $256K$ users with $580K$ unique items. 

\subsection{Attacker Budget}\label{budget_exp}
\begin{figure}[h]
    \begin{center}
        \vskip -0.1in
        \subfigure[Instacart Dataset]{
        \includegraphics[width=\linewidth]{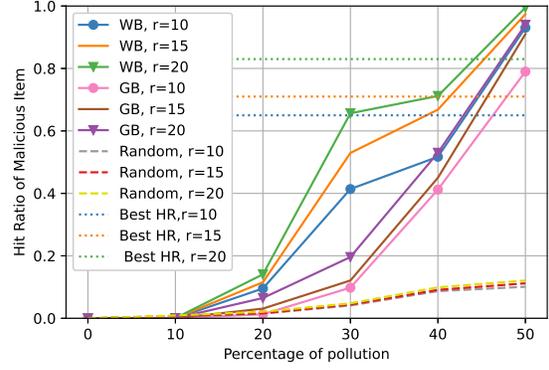}
        \label{hit_ratio:a}}
        \vskip -0.1in
        \subfigure[Proprietary Dataset]{
        \includegraphics[width=\linewidth]{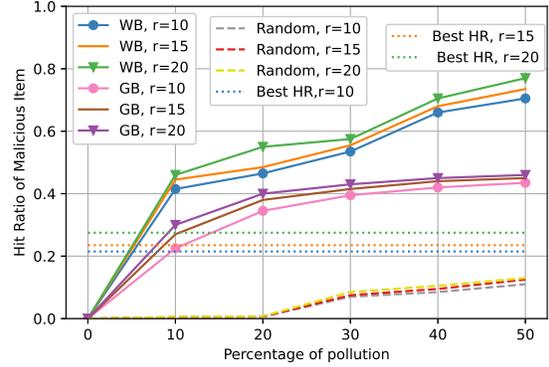}
        \label{hit_ratio:b}}
    \end{center}
    \vskip -0.1in
    \caption{\textbf{Variation of hit-ratio with attacker budget}}
    \label{fig:budget_plot}
\end{figure}

The first constraint introduced in Equation \ref{eq4} is the attacker budget which is the number of malicious users that can be injected into the system. Presented in Figure \ref{fig:budget_plot} is the behavior of hit-ratio with increasing attacker budget.


The behavior of hit-ratio $h_t$ is a function of the number of recommendations $r$ shown to a given user. The hit-ratio is plotted for three values of $r$. As a lower bound, we simulated a random attack of sessions containing the target item $t$ and plotted the resulting hit-ratio for the three values of $r$. For reference, the value of the best hit-ratio in the system for any item, prior to the pollution attack is plotted as well. 
It is observed that the value of the hit-ratio increases with increasing budget across both datasets. Also, both the gray-box and white-box attacks are significantly ahead of the random attack.

In the Instacart Dataset, due to repeat purchase behavior of users on grocery items, strong associations are formed and the best hit-ratio item prior to the attacks has a value ranging from 0.6 to 0.85 for the values of $r$. 
For the proprietary dataset, to maintain confidentiality, we only report scaled versions of the observed hit-ratio.  The number of items were much larger and thus item-item associations were rather weak resulting in pretty low values for the best hit-ratio items. 
For the gray-box attack, the percentage $p$ of data visibility is an important parameter. At $p=0$, the attack would be completely random. At $p=100$, the attack would be a white-box attack. In our experiments we found that for values of $p$ less than 10, the generated recommender system from the warm up phase is close to random. The results plotted are for a value of $p=10$ (since we are attempting to reveal as less information to the attacker as possible).

Another parameter to be tuned for the gray-box attack was the warm up ratio $\epsilon$ used to generate the approximation for the recommender system. Since the attacker wants to use as less budget as possible for the warm up phase, the minimum value of $\epsilon$ for which an approximation of the original Recommender System could be generated is needed. The best value of $\epsilon$, in our experiments was 0.20 for both the datasets. 
\subsection{Data Pollution Attack Detection}
 In this section, we simulate both white-box and gray-box data pollution attack schemes. Per attack scenario the attacker injected $|U^m|$ users, where each injected user generates $o$ sessions. 
\subsubsection{Baseline detection}
To put our results into perspective, we developed a baseline for detecting the pollution sequences. The baseline is an LSTM which is trained on heuristically clean parts of the data as real data and random sequences as negatives (similar to the pretraining step of the generator in our sequential GAN architecture). The decision threshold for the LSTM is the 99th percentile of negative log-likelihood (NLL) scores on a cross-validation set.
Once the LSTM is trained, we evaluate sequences by applying the threshold on the NLL score of those sequences. A low NLL score is expected to be a real sequence and higher NLLs will indicate polluted sequences. It is worth noting that while this baseline approach performs really well in detecting random attacks, a well disguised attack can easily fool this detection algorithm. 
\subsubsection{Attack detection performance}
\begin{figure}[ht]
    \begin{center}
    \vskip -0.1in
    \subfigure[Attack detection on Instacart Dataset]{
        \includegraphics[width=\linewidth]{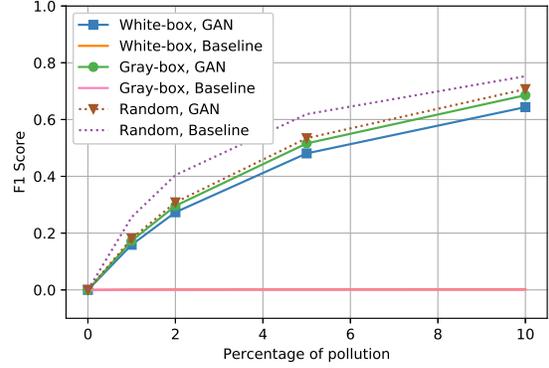}
        \label{detection:a}}
    \vskip -0.1in
    \subfigure[Attack detection on proprietary Dataset]{
        \includegraphics[width=\linewidth]{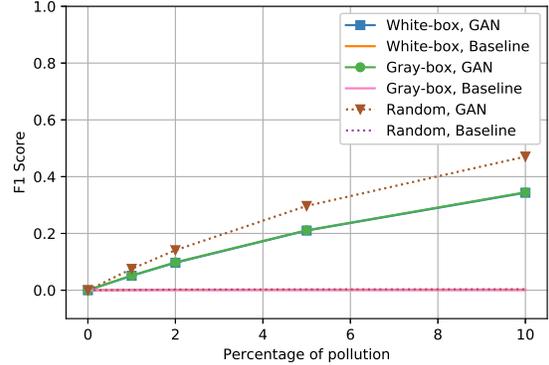}
        \label{detection:b}}
    \caption{\textbf{Attack detection performance of the proposed sequential GAN algorithm}}
    \label{detection_fig}
    \end{center}
\end{figure}

The results of our proposed sequential GAN architecture as well as the baseline detection are presented in Figure \ref{detection_fig}. The Panels of this Figure represent the performance of proposed sequential GAN and the baseline detection under white-box attack (Algorithm \ref{white_box_algo}), gray-box attack (Algorithm \ref{gray_box_algo}), and random attacks on both Instacart and the proprietary datasets. 

The $x$-axis in Figure \ref{detection_fig} represents the attacker's resources in terms of percentage injected sessions $o|U^m|$ to the total number of user sessions. The $y$-axis represents F1 score for detecting the pollution sequences.
In both Figures \ref{detection_fig}a) and \ref{detection_fig}b) the F1 score is growing with the attacker adds more pollution sequences. This is driven by the increasing precision of the algorithm as number of attack sequences grow. In addition, on both datasets the baseline detection has precision and recall very close to zero for the gray-box and white-box attacks. This clearly shows that that the baseline will completely fail in detecting well-disguised attacks. while the proposed sequential GAN detection performs significantly better.

Another observation is that the performance of the proposed approach is better on the Instacart dataset. The reason is that the proprietary dataset is sparser and has fewer sessions per item on average. Hence the GAN architecture does not find as many patterns as it would in the case of Instacart dataset which is more dense in comparison (The proprietary dataset has $1.85K$ sessions per $580K$ items while Instacart has $3.6M$ session for $206K$ items). For the same reason, the performance of baseline approach on random attacks is completely different for the two datasets. In case of the proprietary dataset, the sequences look more random-like therefore the NLL score for real sequences and pollution sequences are very close while in the case of Instacart there is a significant gap in NLL scores of the real sequences vs random ones. 
\vspace*{-3 mm}
\section{CONCLUSION}
In this work, we presented a GAN-based algorithm to detect the injection of data pollution sequences into the recommender system under two attack scenarios: white-box and gray-box. The proposed algorithm is able to efficiently take into account the context of user sessions by applying a language model on contextual features of items. The proposed context-aware sequential GAN performs significantly better than the LSTM based baseline approach on two real world e-commerce datasets. Implementation of such detection algorithms will drastically diminish any gain that an attacker hopes to achieve by launching the pollution attack. 
\bibliography{sample-base}
\bibliographystyle{icml2020}
\end{document}